\documentclass[letterpaper, 10 pt, conference]{ieeeconf}  
\IEEEoverridecommandlockouts                              
\usepackage{times}
\usepackage[pdftex]{graphicx}
\usepackage{amsmath,amsfonts,amssymb}
\usepackage{graphicx}
\usepackage{prettyref}
\usepackage{mathrsfs}
\usepackage{wrapfig}
\usepackage{MnSymbol}
\usepackage{multirow}
\usepackage{xcolor}
\usepackage{booktabs}
\usepackage{algorithm,algpseudocode}

\newrefformat{Fig}{Figure~\ref{#1}}
\newrefformat{Appen}{Appendix~\ref{#1}}
\newrefformat{Sec}{Section~\ref{#1}}
\newrefformat{Table}{Table~\ref{#1}}
\newrefformat{Alg}{Algorithm~\ref{#1}}
\newrefformat{Line}{Line~\ref{#1}}
\newrefformat{Def}{Definition~\ref{#1}}
\newrefformat{Thm}{Theorem~\ref{#1}}
\newrefformat{Step}{Step~\ref{#1}}
\newrefformat{Pb}{Problem~\ref{#1}}
\newrefformat{Eq}{Equation~\ref{#1}}

\newcommand{\E}[1]{\mathbf{#1}}

\newcommand{\TWOC}[2]{\left(\begin{array}{c}#1 \\ #2\end{array}\right)}

\newcommand{\THREEC}[3]{\left(\begin{array}{c}#1 \\ #2 \\ #3\end{array}\right)}

\newcommand{\MTTTHREE}[9]{\left(\begin{array}{ccc}#1 & #2 & #3 \\ #4 & #5 & #6 \\ #7 & #8 & #9\end{array}\right)}

\usepackage{tikz}
\usetikzlibrary[arrows,backgrounds,decorations.pathmorphing,positioning,fit,petri]

\newcommand{\RC}{\mathcal{C}}

\newcommand{\LC}{\mathcal{L}}

\newcommand{\DM}[1]{}
\newcommand{\COMM}[2]{}{}
\newcommand{\FINE}[1]{}
\algnewcommand{\LineComment}[1]{\State \(\triangleright\) #1}

\title{\LARGE \bf Motion Planning for Fluid Manipulation using Simplified Dynamics}
\author{Zherong Pan$^{1}$ and Dinesh Manocha$^{2}$  \\
\small{video: https://www.dropbox.com/sh/b4wd95uob8tfc6q/AACTGK-CAaXTjj8rHgdIiI8Ga?dl=0}
\thanks{$^{1,2}$Zherong and Dinesh are with Department of Computer Science, the University of North Carolina,
        {\tt\small \{zherong,dm\}@cs.unc.edu}}%
}

\begin{document}
\maketitle
\thispagestyle{empty}
\pagestyle{empty}

\begin{abstract}
We present an optimization-based motion planning algorithm to compute a smooth, collision-free trajectory for a manipulator used to transfer a liquid from a source to a target container. We take into account fluid dynamics constraints as part of trajectory computation. In order to avoid the high complexity of exact fluid simulation, we introduce a simplified dynamics model based on physically inspired approximations and system identification. Our optimization approach can incorporate various other constraints such as collision avoidance with the obstacles, kinematic and dynamics constraints of the manipulator, and fluid dynamics characteristics. We demonstrate the performance of our planner on different benchmarks corresponding to various obstacles and container shapes. Furthermore, we also evaluate its accuracy by validating the motion plan using an accurate but computationally costly Navier-Stokes fluid simulation.
\end{abstract}
\section{INTRODUCTION}
Motion planning is a core problem in robotics and has been studied for many decades. It is typically used to compute a collision-free path for moving an object from an initial position to a goal position or for assembly tasks. Some of the earlier applications of motion planning were restricted to the handling of rigid or articulated models. Recently, there has been interest in manipulating highly deformable objects such as cloth \cite{li2015regrasping}, ropes \cite{huang2015leveraging}, or human tissues \cite{chentanez2009interactive}, which can have higher physical-simulation complexity. In this paper, we deal with planning for fluid manipulation tasks performed using a robot, including liquid transfer where the main goal is to transfer a liquid from one container to the other. This problem is important in industrial applications, where the robots are used to transfer dangerous fluids or chemicals, as part of handling materials, cleaning tasks, or using lubricants. Other applications arise in domestic and service tasks, where robots may be used to refill a cup of coffee or to pour liquids from a bottle to a glass. Humans tend to be quite good at learning fluid manipulation tasks quickly and can easily exploit the physical properties of the fluid. However, it is non-trivial for the robot to perform such tasks due to the intrinsic challenges of modeling the fluid dynamics.

\begin{figure}
\begin{center}
\includegraphics[trim=0mm 0mm 0mm 0mm, clip, width=.99\linewidth]{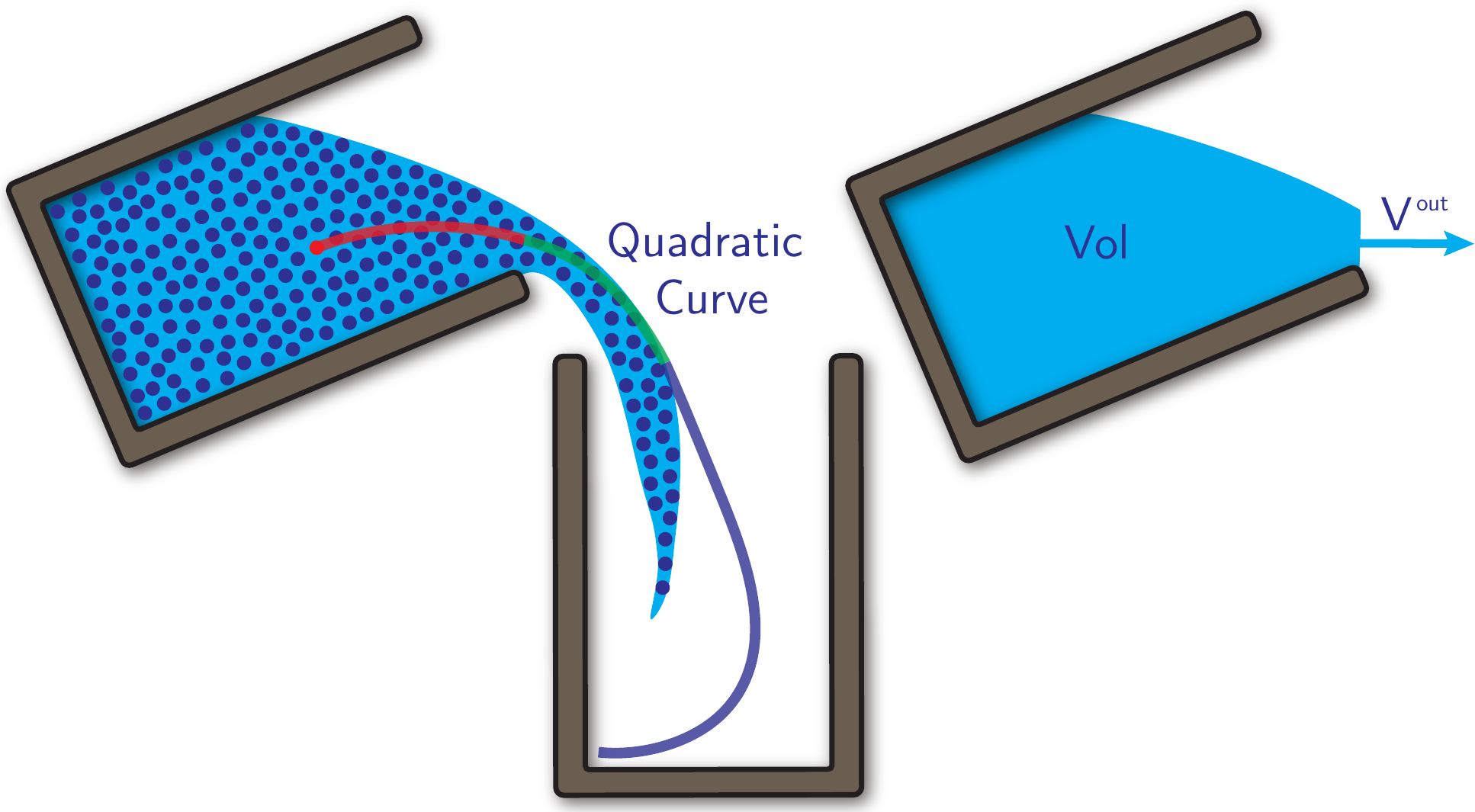}
\put (-195,55) {(a)}
\put (-55,55) {(b)}
\end{center}
\vspace{-10pt}
\caption{\label{Fig:FluidRep} An illustration of the simplified fluid body representation. (a): Conventionally, a large set of parameters are needed to represent fluid bodies, e.g., by a set of particles. We assume that each fluid particle's streamline is divided into three stages: During the source stage (red) liquid particle lies within the container. During the free flying stage (green), liquid particle leaves the container and follows a quadratic curve $\E{C}$. Finally, the target stage (blue) begins when the particle falls into $\E{T}$. (b): Our simplified dynamics model parametrizes fluid body using only two global descriptors: the volume of fluid in the source container $vol$ (blue region), and the mean outflowing velocity magnitude $v^{out}$.}
\vspace{-10pt}
\end{figure}

There are many issues that arise in designing such motion planning algorithms. The first issue arises due to the flexibility of fluid body. Fluid body can undergo complex topology changes and large deformations as shown in \prettyref{Fig:FluidRep}. As a result, we need to use appropriate data structures that can provide flexibility, e.g., using a large set of particles. Unfortunately, these representations, usually parametrize the fluid body using tens of thousands of variables. It is difficult for any existing motion planning algorithm to take into account deforming obstacles specified using so many degrees of freedom. Moreover, fluid body dynamics is governed by the Navier-Stokes equation, which is a nonlinear PDE. In order to model accurate fluid manipulation for planning, this PDE has to be used to specify appropriate constraints, and solved during each step of the planning to compute a valid configuration. However, exact solvers for the Navier-Stokes equation can be expansive \cite{Zherong:2016:ICAPS}.

\begin{figure}
\vspace{-10pt}
\begin{center}
\setlength{\tabcolsep}{1pt}
\begin{tabular}{ll}
\includegraphics[trim=0mm 0mm 0mm 0mm, clip, width=.8\linewidth]{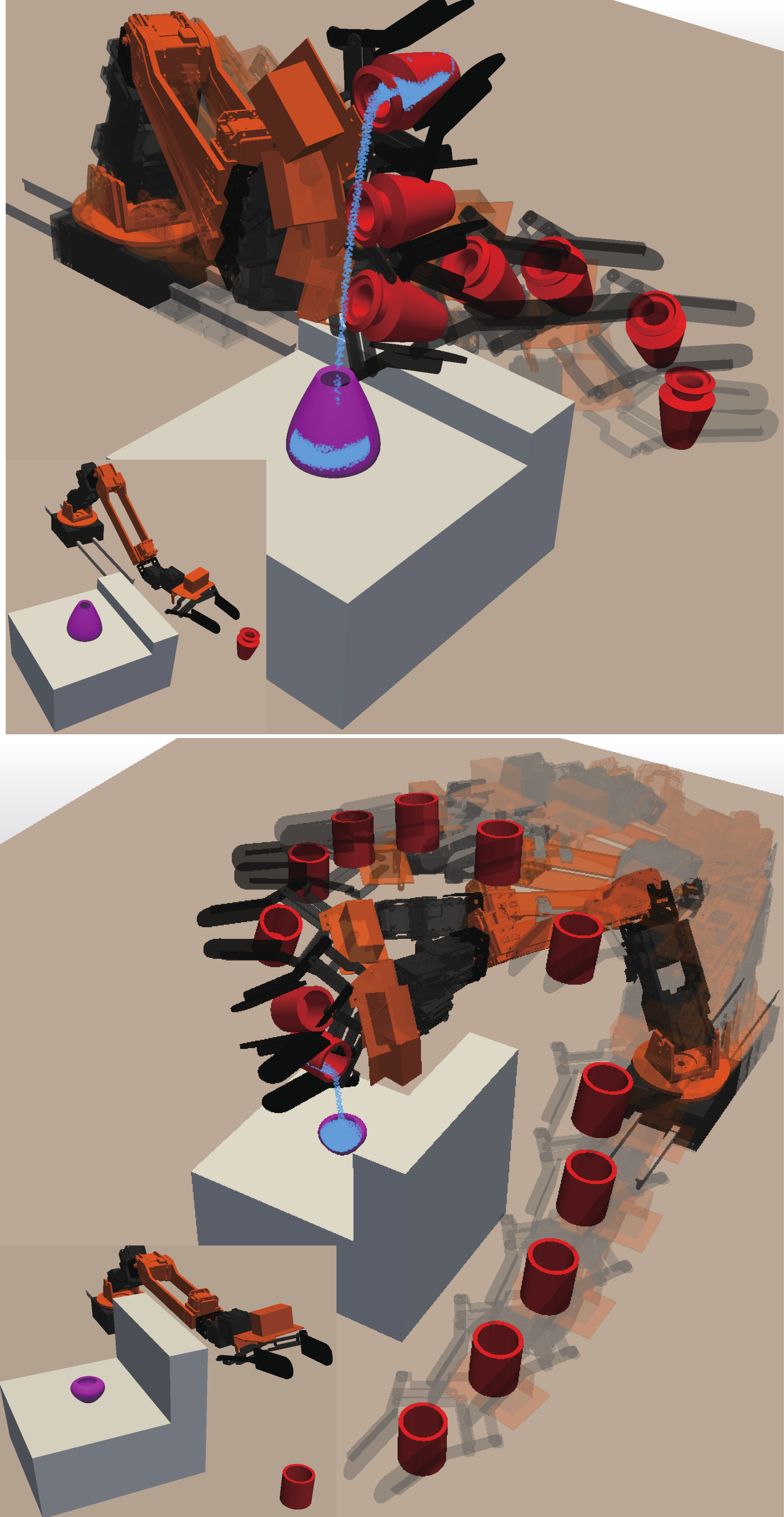}
\end{tabular}
\put (-195,70) {(a)}
\put (-195,185) {(b)}
\put (-195,-130) {(c)}
\put (-195,0) {(d)}
\end{center}
\vspace{-10pt}
\caption{\label{Fig:Bench} We highlight some of the challenging environments used to evaluate the perofrmance of our motion planner.   (a): Guided by a simplified dynamics model, our planner can predict the liquid outflow curve. This enables us to successfully transfer liquid even when the source container is far away from the target container, as shown in (b). (c): Our optimization based motion planner can compute collision-free paths, while performing fluid manipulation and transfer tasks, e.g. a high block is an obstacle between the source and the target container, as shown in (d).}
\vspace{-20pt}
\end{figure}
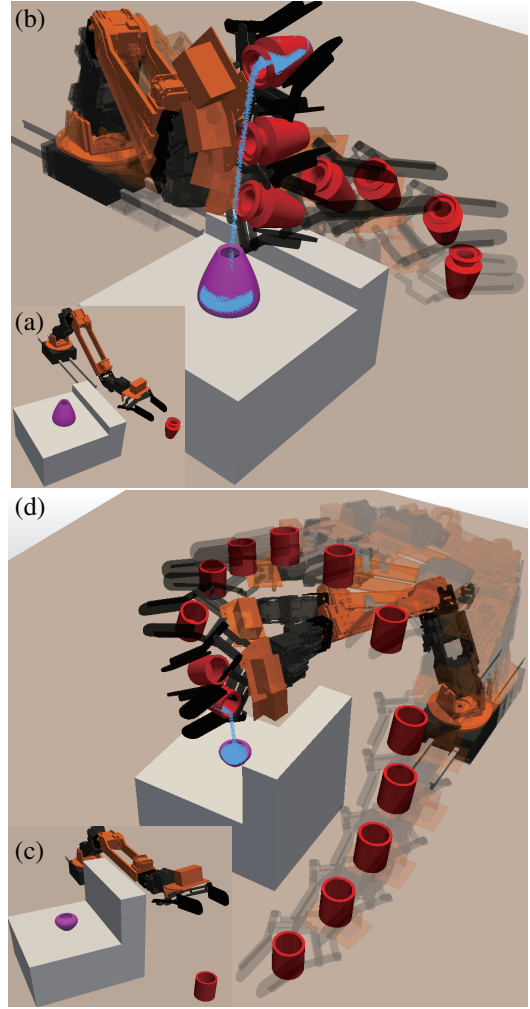
{\bf Main Results:} In this paper, we introduce a simplified fluid dynamics model, which can be specified by relatively few parameters that involve only global descriptors: the total volume of the fluid in the container, and the outflowing velocity magnitude. In this parameter space, we use physically inspired approximations and system identification techniques to derive the governing equations for the global descriptors. As illustrated in \prettyref{Fig:FluidRep}, this approximation is based on the observation that, due to the smoothness requirements placed on the trajectories, the liquid outflow from the container has strong regularity and it approximately follows a quadratic curve. This quadratic curve can be parametrized only by the velocity magnitude and direction, in the container's local coordinates.

We combine this simplified dynamics model with an optimization-based planner that can compute collision-free and smooth trajectories, and also satisfy the fluid dynamics constraints. Our planner based on the simplified dynamics model can achieve more than two orders of magnitude speedup compared with one that uses an accurate solver. It can also be easily integrated into current optimization-based planning approaches \cite{Ratliff:2009,STOMP:2011,schulman2014motion}. In \prettyref{Fig:Bench}, we demonstrate the performance of our method based on using different benchmarks with different shape of obstacles between the source and target containers. Our simplified model can reduced the planning time from hours to minutes.

The rest of the paper is organized as follows. After reviewing some related works in \prettyref{Sec:RW}, we give an overview of our planning algorithm in \prettyref{Sec:OV}. We then present the simplified dynamics model in \prettyref{Sec:SDM}. Finally, we evaluate the accuracy of our simplified model and validate our motion plan by comparing with results from accurate Navier-Stoke's-based simulation in \prettyref{Sec:ER}.

\section{\label{Sec:RW}RELATED WORKS}
In this section, we give a brief overview of prior works on motion planning, fluid dynamics, and planning with fluid constraints.

\subsection{GENERAL MOTION PLANNING}
Robotic motion planning can always be formulated as a trajectory search problem under various constraints. Early works of motion planning \cite{lavalle1998rapidly} focus solely on feasibility under simple collision-free constraints. Many extensions have been proposed in \cite{GRVO:2009,stilman2007task,berenson2009manipulation} to handle other types of constraints. Some of the widely used algorithms for motion planning are sampling-based algorithms, but it is hard to integrate some kinds of constraint forms into those methods. Recent works focus more on computing locally optimal motion plans, see e.g. \cite{Ratliff:2009,STOMP:2011,Park:2012:ICAPS}. Based on general-purpose numerical optimization methods, these planners have the advantage in that they can naturally account for various additional requirements, such as smoothness, collision-avoidance, etc. This property makes them especially attractive for object manipulation tasks, including fluid manipulation. Specifically, our planning framework uses the optimization approach described in \cite{schulman2014motion}, which is an optimization algorithm that can exactly satisfy the collision constraints. 

\subsection{FLUID DYNAMICS AND PLANNING}
In order for a robot to interact with dynamics environmental objects such as fluids, it is important to estimate and predict their dynamics states. A well-established way for such estimation is based on computational fluid dynamics (CFD)  \cite{anderson1995computational}. CFD simulators can be categorized by the different ways used to represent the fluid body: meshless methods \cite{zhu2005animating,barreiro2013smoothed} represent fluids using a set of particles; grid based methods \cite{harlow1965numerical,osher2006level} represent it as material flowing through static spatial cells; FEM methods \cite{Clausen:2013:SLS} instead use spatial cells that move with fluid bodies. To represent a small block of water, e.g. a cup of coffee in a mug, all these methods tend to use tens of thousands of parameters, taking hours for even one pass of forward simulation on a desktop PC. This makes it almost impractical to perform a numerical search for a solution in such high dimensional spaces. Indeed, earlier works like \cite{mcnamara2004fluid} take days to control a short fluid simulation corresponding to an animation sequence.

\subsection{PLANNING WITH FLUID CONSTRAINTS}
There are some prior works on motion planning for fluid manipulation. We classify these works into three categories: methods using demonstration and machine learning \cite{langsfeld2014incorporating,lee2014unifying,alatartsevrobot,yamaguchi2015pouring}, methods based on full-featured dynamics model, (i.e. CFD simulator) \cite{kuriyama2008trajectory,Zherong:2016:ICAPS}, and methods using simplified dynamics model \cite{kunze2011simulation,tzamtzi2008robustness}. Machine learning based methods work around the problem of high dimensionality by generalizing from a set of already optimized, learned, or captured ``example'' trajectories. However, obtaining such example trajectories for fluid manipulation is usually non-trivial. Moreover, these methods have difficulties generalizing to new scenarios, as new datasets are needed. In order to handle new scenarios automatically, some methods use an accurate full-featured CFD simulator. \cite{kuriyama2008trajectory} uses evolutionary optimization and \cite{Zherong:2016:ICAPS} uses an EM-like optimization algorithm, first assuming fixed fluid particle streamlines within each iteration and then correcting those streamlines using an additional forward CFD simulation. Both these methods have low performance since they require multiple passes of costly fluid simulations. In view of this high complexity, \cite{kunze2011simulation,tzamtzi2008robustness} use a reduced or simplified dynamics fluid model to achieve higher planning performance.
\section{\label{Sec:OV}OVERVIEW}
In this section, we formulate our problem, introduce the notation, and give an overview of our approach.

\subsection{PROBLEM DEFINITION}
\begin{figure}
\begin{center}
\includegraphics[trim=0mm 0mm 0mm 0mm, clip, width=.95\linewidth]{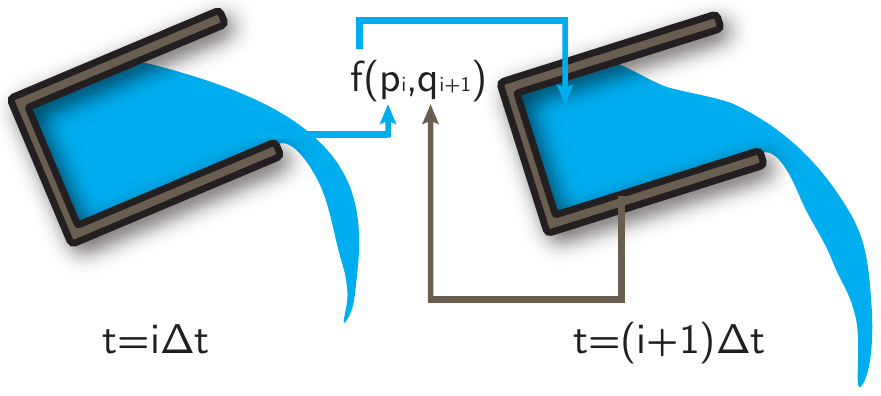}
\end{center}
\vspace{-10pt}
\caption{\label{Fig:Evolution} An illustration of augmented parameter space of our fluid model and the governing PDE that highlights the relationship between the parameters of consecutive timesteps.}
\vspace{-10pt}
\end{figure}

In order to deal with a coupled planning problem that handles the constraints of the robot and the fluid body, we use the notion of an augmented parameter space: $\RC\times\LC$. Here $\RC$ is the robot's kinematic configuration space: Each point $\E{q}\in\RC$ consists of all the joint variables characterizing the position of each rigid link. $\LC$ is the set of parameters representing a fluid's dynamics state: each point $\E{p}\in\LC$ consists of all the variables characterizing the position as well as the velocity of the fluid body. We formulate the motion planning problem as an optimization over the space of valid trajectories $<\mathcal{Q}(t),\mathcal{P}(t)>$. Here $\mathcal{Q}(t),\mathcal{P}(t)$ are robot and fluid trajectories respectively, functions mapping $t\in[0,\tau]$ to $\RC,\LC$ ($\tau$ is the duration of the trajectory). Both these trajectory functions are discretized using finite difference with timestep size $\Delta t=\tau/(N-1)$.
\begin{eqnarray*}
\E{Q}=\left[\E{q}_1^T,\cdots,\E{q}_N^T\right]^T  \quad \E{P}=\left[\E{p}_1^T,\cdots,\E{p}_N^T\right]^T,
\end{eqnarray*}
are the set of discrete trajectory samples where $\E{q}_i=\mathcal{Q}((i-1)\Delta t),\E{p}_i=\mathcal{P}((i-1)\Delta t)$. In order to model the dynamics of environmental objects, consecutive trajectory samples $\E{p}_{i,i+1}$ are related by additional governing PDEs and spatial/temporal boundary conditions, e.g. the Navier-Stokes equation is used in \cite{Zherong:2016:ICAPS}. We can encode this relationship as a function: 
\begin{eqnarray}
\label{Eq:PDE}
\E{p}_{i+1}=f(\E{p}_i,\E{q}_{i+1}), 
\end{eqnarray}
where $f$ takes the dynamics state $\E{p}_i$ at last timestep and the boundary condition $\E{q}_{i+1}$ at current timestep and integrates the governing PDE over time $\Delta t$ to compute the dynamics state at the current timestep $\E{p}_{i+1}$ as illustrated in \prettyref{Fig:Evolution}. Other planning problems with dynamics environmental objects can also fit in such a framework, but we use a different function $f$ to represent the physics constraints. A common assumption taken by previous works \cite{frank2014learning,lee2014unifying} is that the magnitude of end-effector velocity is much smaller than the characteristic rate of energy dissipation due to friction or damping, so that dependencies between consecutive timesteps can be ignored. And we can use a simplified relationship $\E{p}_{i+1}=f(\E{q}_{i+1})$. Since most fluid materials have a low energy dissipation rate (i.e. small viscosity), we can't use this assumption in our formulation.

\subsection{COMPLEXITY OF FLUID MANIPULATION}
$\LC$ is usually of a much higher dimension than $\RC$. If we use a particle based representation as illustrated in \prettyref{Fig:FluidRep}, each particle introduces an additional $6$ dimensions (3 for position and 3 for velocity) to $\LC$. As a result, time integration of the governing PDE over a small timestep $\Delta t$ can take minutes, and updating the whole $\E{P}$ will therefore take hours on a desktop computer. Furthermore, an optimization algorithm needs to perform multiple updates of $\E{P}$. In order to overcome this complexity, we use a simplified dynamics model for fluid manipulation.

\subsection{OPTIMIZATION PIPELINE}
Given the specification of the parameters and the (discrete) trajectory space, \cite{schulman2014motion} plans the motion of the robot using numerical spacetime optimization over the vector space $<\E{Q},\E{P}>$ under various non-linear constraints. It can be posed as:
\begin{eqnarray*}
\E{argmin}_{\E{V}}&&C_{obj}(\E{Q},\E{P}) + C_{reg}(\E{Q}) \\
\E{s.t.}&&C_{coll}(\E{Q}) \geq 0 \quad C_{robot}(\E{Q}) =/\geq 0    \\
        &&C_{pde}(\E{Q},\E{P}) = 0.
\end{eqnarray*}
We briefly review the terms used in our planning algorithm. $C_{coll}$ requires each $\E{q}_i$ to lie in $\RC_{free}$ (the subspace of $\RC$ where no collision occurs between the robot and obstacles). To satisfy this requirement, we perform discrete collision checking between all pairs of objects and robot links at each timestep $i$ to find a set of $K_i$ invalid penetrations $\{a^j(\E{q}_i),b^j(\E{q}_i),n^j,d^j|j\in 1,\cdots,K_i\}$, where $a^j(\E{q}_i),b^j(\E{q}_i)$ are a pair of points in contact, $n^j$ is the contact normal and $d^j$ is the penetration depth. These constraints are imposed by setting $C_{coll}=<n^j,a^j(\E{q}_i)-b^j(\E{q}_i)>-d^j$. In order to reduce the problem size, we only retain the deepest pair of penetrations for each pair of colliding objects. $C_{robot}$ is the set of inherent constraints in the robot, like the kinematic and dynamics constraints. In this work, we only consider the joint limits and the maximum velocity constraints. Finally, $C_{pde}$ corresponds to the constraints that the relationship in \prettyref{Eq:PDE} is satisfied exactly. $C_{reg}$ corresponds to the smoothness characteristics of the trajectory space. We use the formulation in \cite{STOMP:2011,Park:2012:ICAPS} and use a Laplacian penalty term in the joint space:
\begin{eqnarray*}
C_{reg}=\frac{1}{2}\Sigma_i\|\E{q}_{i-1}-2\E{q}_i+\E{q}_{i+1}\|^2.
\end{eqnarray*}
Finally, $C_{obj}$ encodes the high level goals of transferring liquid from a source to a target container. We give its definition in the next section. With a smooth and differentiable $C_{obj},C_{pde}$, this optimization can be performed using local linearization. However, this may not be the case with fluids, where $f$ and $C_{pde}$ are usually non-smooth even for our simplified model. Therefore, we follow the approach in \cite{Zherong:2016:ICAPS} and decouple the planning problem: first fixing $\E{P}$ and updating $\E{Q}$ by solving the subproblem:
\begin{eqnarray*}
\E{argmin}_{\E{V}}&&C_{obj}(\E{Q}) + C_{reg}(\E{Q}) \\
\E{s.t.}&&C_{coll}(\E{Q}) \geq 0 \quad C_{robot}(\E{Q}) =/\geq 0.
\end{eqnarray*}
We then update $\E{P}$ by applying \prettyref{Eq:PDE} repeatedly.

In order to solve this sub-problem numerically, we use the sequential quadratic programming (SQP) with a trust region algorithm in the outer loop. The trust region is adjusted using the Levenberg-Marquardt algorithm according to \cite{nielsen1999damping}. Special care is needed in the inner loop. Since conflicting constraints may be returned by collision checking, directly linearizing each collision constraint $C_{coll}$ will sometimes result in an infeasible QP problem. Therefore, we impose $C_{coll}$ as a soft penalty:
\begin{align}
\label{Eq:CollCons}
\begin{split}
C_{collP}^1=\eta \Sigma_i\E{t}_i \quad\E{s.t.}\quad \E{t}\geq0\quad \E{t}\geq -C_{coll} \\
C_{collP}^2=\mu\|C_{coll}-\E{s}\|_2^2-\lambda^T(C_{coll}-\E{s}) \quad\E{s.t.}\quad \E{s}\geq0 ,
\end{split}
\end{align}
where the first formulation is the L1-penalty used in \cite{schulman2014motion} and the second formulation is the Augmented Lagrangian-penalty. Either formulation requires additional parameters $\eta$ or $\mu,\lambda$ to be adjusted according to \cite{nocedal2006numerical}. The final optimization pipeline is summarized in \prettyref{Alg:pipeline}. Contrasted to \cite{Zherong:2016:ICAPS} where $\E{P}$ is updated in the outer loop (before \prettyref{Line:Outer}), we can update $\E{P}$ more frequently in the inner loop (\prettyref{Line:Inner}), due to the lower computational overhead of the simplified dynamics model.

\begin{algorithm}[h]
\caption{Spacetime Optimization Algorithm}
\label{Alg:pipeline}
\begin{algorithmic}[1]
\Require Robot/Environment, $C_{pde},C_{obj}$
\Require Initial $k,\eta,\mu,\lambda,\E{Q}_0,\E{P}_0$
\Ensure Locally optimal $\E{Q,P}$
\While{$k=1,\cdots$}
\State Do collision checking to find $C_{coll}$
\State Set {$\E{Q}_{k}=\E{Q}_{k-1}$}

\While{true}\label{Line:Outer}
\LineComment Update $\E{P}$ from $\E{Q}$
\State Compute $\E{P}_k$ by applying \prettyref{Eq:PDE}    \label{Line:Inner}
\LineComment Define the QP problem objective at $\E{Q}_k$
\State Local quadratic expansion of objective
\State \;\;$C_{obj}+C_{reg}+C_{collP}^*(\eta,\mu,\lambda)\approx$
\State \;\;$\frac{1}{2}\E{Q}^T\E{H}\E{Q}+\E{b}^T\E{Q}+c$
\State Apply trust region by setting $\E{H}=\E{H}+k\E{I}$
\LineComment Define the QP problem constraints at $\E{Q}_k$
\State Add constraints for $\E{t},\E{s}$ (\prettyref{Eq:CollCons})
\LineComment Robot kinematic, dynamics constraints
\State Local linear expansion of $C_{robot}$
\State Solve semi-definite QP for $\E{Q}^*$
\If{$\|\E{Q}^*-\E{Q}_{k}\|_0 < \epsilon$}
\State Break
\EndIf

\State Set $\E{Q}_k=\E{Q}^*$
\LineComment Update collision penalty stiffness
\If{$C_{collP}^*=C_{collP}^1$}
\State Update $\eta$ (\cite{nocedal2006numerical})
\EndIf
\If{$C_{collP}^*=C_{collP}^2$}
\State Update $\mu,\lambda$ (\cite{nocedal2006numerical})
\EndIf
\EndWhile

\If{$\|\E{Q}_{k}-\E{Q}_{k-1}\|_0 < \epsilon$}
\State Break
\EndIf
\LineComment Update trust region
\State Update $k$ (\cite{nielsen1999damping})
\EndWhile
\end{algorithmic}
\end{algorithm}
\section{\label{Sec:SDM}SIMPLIFIED DYNAMICS MODEL}
The goal of liquid transfer is to move liquid material from a source container $\E{S}$ to a target container $\E{T}$. In this work, we make several assumptions: the source container is assumed to be an axial symmetric rigid body, the target container $\E{T}$ is fixed and the source container $\E{S}$ is attached to the end-effector of a robot arm. In this section, we derive the formulation of our simplified governing PDE $f$ and the objective function $C_{obj}$. Plugging these definitions into \prettyref{Alg:pipeline} corresponds to the overall planning algorithm. 

A spectrum of governing PDEs have been developed to model the motion of a fluid body, working under various assumptions and data-structures, see e.g. \cite{anderson1995computational}. The choice of appropriate governing PDE is critical to the success of a liquid transfer planner. The most general governing PDE for the fluid body is the Navier-Stokes equation. Although a direct discretization of this equation is also possible, it introduces a large number of parameters to represent the dynamics state of fluid body. In view of this, previous work \cite{Zherong:2016:ICAPS} assumes the streamline of each fluid particle doesn't change within the inner loop of \prettyref{Alg:pipeline}. As a result, a forward simulation in the outer loop is needed to correct the streamline. This method simplifies the optimization problem but the repeated costly forward simulations in the outer loop considerably slow down the overall performance. With these limitations, most planning algorithms instead use a simplified or reduced fluid models, e.g., the pendulum model \cite{tzamtzi2008robustness}. Although this is easy to implement, cheap to compute and accurate when $\E{S}$ is undergoing only a slight perturbation, this method cannot account for significant container movements, e.g. when pouring the liquid from $\E{S}$ to $\E{T}$.

In this paper, we present a simplified dynamics model designed exclusively for the liquid transfer tasks. Our design guideline is to reduce the number of parameters as much as possible, while at the same time to retain the ability of predicting the locus of fluid particles that leave the container $\E{S}$. In addition, we want to account for variations in the environment, such as different container shapes and different liquid materials. 

As illustrated in \prettyref{Fig:FluidRep}, we first simplify the problem by dividing the streamline of each fluid particle into three stages: during the source stage a fluid particle lies within the bounding volume of $\E{S}$; it then leaves $\E{S}$ and follows a quadratic curve $\E{C}$, which is the free flying stage; finally the target stage begins when it falls into the bounding volume of $\E{T}$. In this work, we ignore the dynamics of the third stage, assuming no spilling after the particle falls into $\E{T}$. As a result, the objective of our planner is to ensure that the quadratic curves $\E{C}$ passes through the center of opening of $\E{T}$. Since $\E{C}$ is a quadratic curve, it is characterized by the outflowing velocity of the fluid and global orientation of $\E{S}$ at each timestep $i$.

\subsection{\label{Sec:GPDE}THE DYNAMICS MODEL}
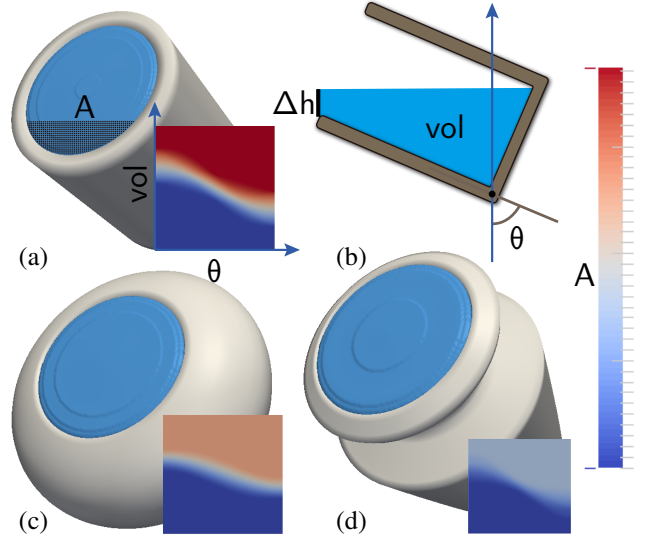
\begin{figure}[t]
\begin{center}
\includegraphics[trim=0mm 0mm 0mm 0mm, clip, width=.98\linewidth]{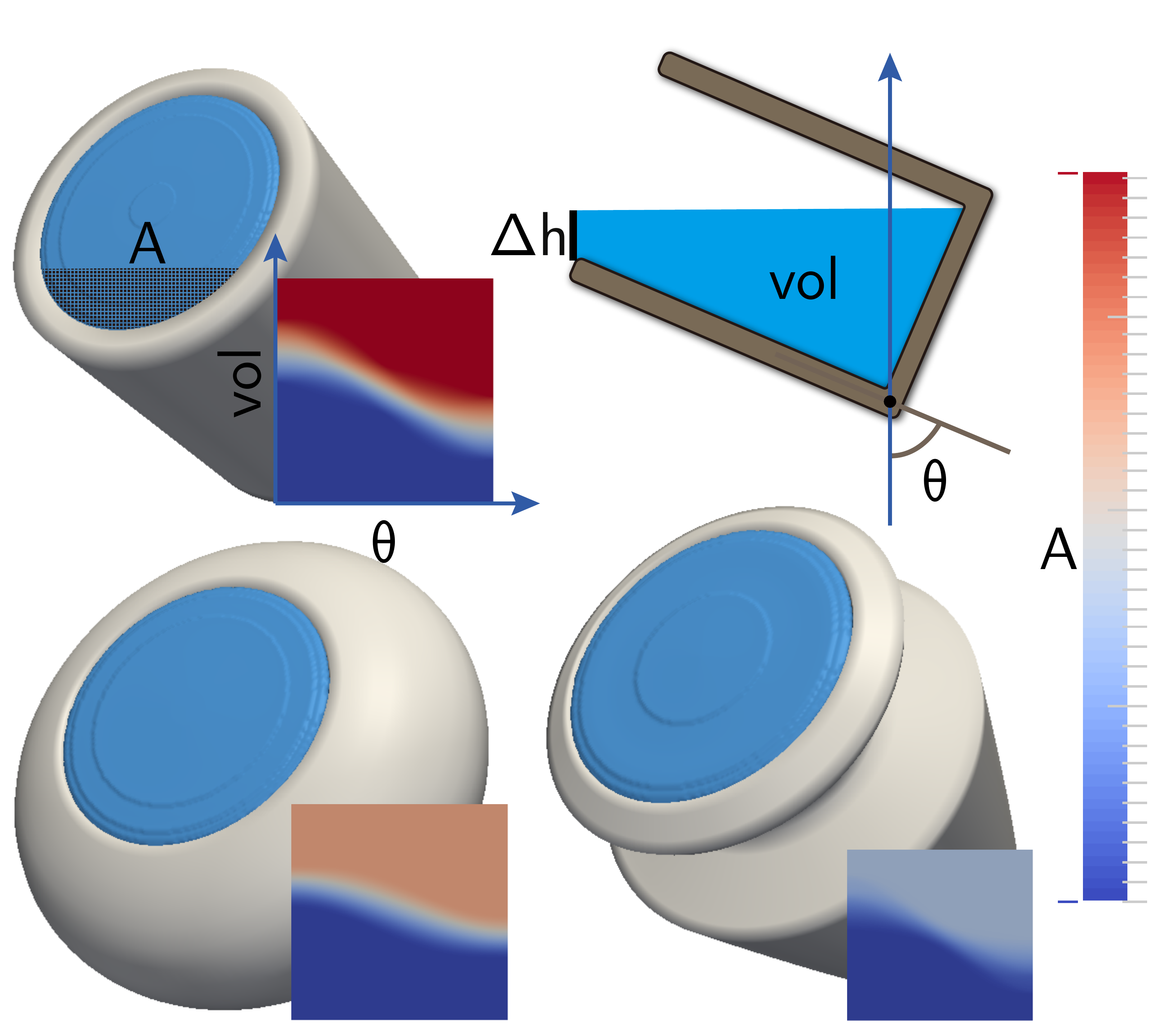}
\put (-235,105) {(a)}
\put (-115,105) {(b)}
\put (-235,5) {(c)}
\put (-115,5) {(d)}
\end{center}
\vspace{-10pt}
\caption{\label{Fig:AFunc} (a): Illustration of function $A(\theta,vol)$. (b): Illustration of function $\Delta h(\theta,vol)$. (c,d): More examples for containers of different shapes. A watershed algorithm is used to extract the interior of $\E{S}$ (blue) and to populate the lookup table of $A(\theta,vol)$ (colormap). The same procedure is used to precompute $\Delta h(\theta,vol)$ on the axial symmetric cross section.}
\vspace{-10pt}
\end{figure}

We first formulate our parameter set. The dimension of $\LC$ is reduced drastically to only two in our model. The components of $\E{p}\in\LC$ are:
\begin{eqnarray*}
\E{p}=\TWOC{vol(m^3)}{v^{out}(m/s)},
\end{eqnarray*}
where $vol$ is the remaining volume of liquid left in the container's bounding volume and $v_{out}$ is the magnitude of the velocity that represents the mean outflow. In order to formulate the timestepping equation of $f(\E{p}_i,\E{q}_{i+1})$, we use forward Euler integration $vol_{i+1}=vol_i-A_{i+1}v_{i+1}^{out}\Delta t$, where $A_{i+1}(m^2)$ is the outflow cross-section area. For a given source container $\E{S}$, we assume $A$ is a function of the leaning angle $\theta$ and volume $vol$, as illustrated in \prettyref{Fig:Local}. This function $A(\theta,vol)$ is precomputed for each given source container $\E{S}$ and stored as a lookup table. To populate the table, we sample $\theta$ at an interval of $1^\circ$, then run a watershed algorithm \cite{roerdink2000watershed} at each $\theta$, while adding an table entry $(A,\theta,vol)$ at each water level. Partial derivatives of this function are approximated using finite differences. An example is given in \prettyref{Fig:AFunc}. Note that the leaning angle $\theta$ is a function of $\E{q}$, which is not a part of $\E{p}$.

\begin{figure}[t]
\begin{center}
\includegraphics[trim=0mm 0mm 0mm 0mm, clip, width=.98\linewidth]{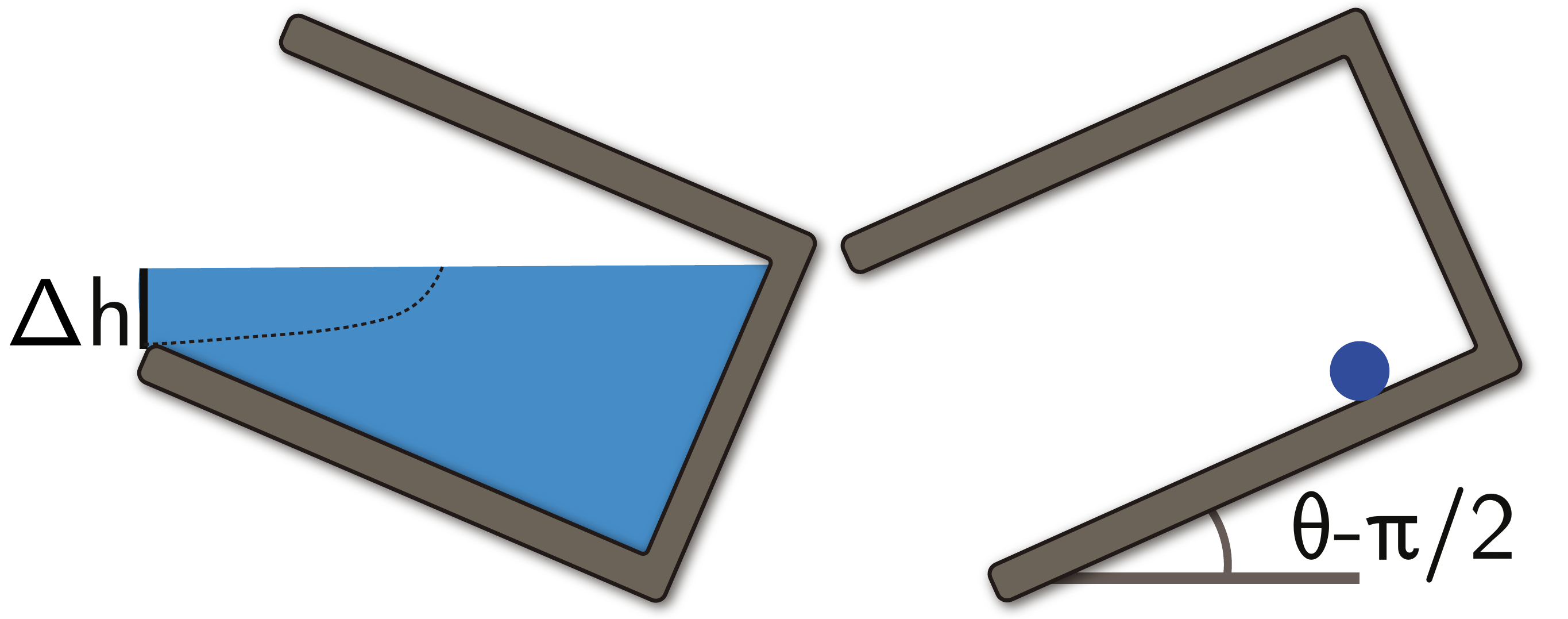}
\put (-235,5) {(a)}
\put (-115,5) {(b)}
\end{center}
\vspace{-10pt}
\caption{\label{Fig:Approx} The two physically inspired approximations. (a): The Bernoulli equation is used between two end points of the dashed streamline. (b): A single fluid particle sliding down the wall of the container.}
\vspace{-10pt}
\end{figure}

Finally in order to determine the magnitude of the mean outflowing velocity $v^{out}$, we combine two physically inspired approximations. The first approximation is based on the Bernoulli equation \cite{bernoulli1738}. This equation states that if the fluid body is steady, inviscid and incompressible, then over each particle streamline, we have: 
\begin{eqnarray*}
p+gh+\frac{v^{out}v^{out}}{2}=\E{constant},
\end{eqnarray*} 
where $p$ is the pressure, $g$ is the gravity and $h$ is the position of the point on the streamline along gravity's direction. Although the assumptions made by the Bernoulli equation are not satisfied exactly, we can still apply it to the two endpoints of the dashed streamline shown in \prettyref{Fig:Approx}. The result implies that $v^{out}$ is related to $\sqrt{2g\Delta h}$. Here $\Delta h$ is a function of $\theta$ and $vol$ computed in the same way as $A(\theta,vol)$, using lookup table. For a 3D source container $\E{S}$, $\Delta h$ is computed for its axial symmetric cross section. The second approximation is based on simple rigid body dynamics. According to \prettyref{Fig:Approx}, when $\theta>\pi/2$, a single particle moving along the wall of the source container $\E{S}$ will gain velocity with magnitude proportional to $\sqrt{sin(\theta-\pi/2)gl}$, where $l$ is the length of the wall. This approximation tells us that $v^{out}$ should be related to $sin(\theta-\pi/2)$. Unfortunately, we don't know exactly what the best relationship model between $v^{out}$ and these two terms is. So that we use system identification to combine their contributions, by making the assumption that:
\begin{small}
\begin{align}
\label{Eq:Model}
\begin{split}
v^{out}\triangleq&g(\theta,vol)=a\sqrt{2g\Delta h}^1+b\sqrt{2g\Delta h}^2+c\sqrt{2g\Delta h}^3+    \\
&d\E{sin}(\E{max}(\theta-\pi/2,0))+e\E{sin}(\E{max}(\theta-\pi/2,0))^2+ \\
&f\E{sin}(\E{max}(\theta-\pi/2,0))^3,
\end{split}
\end{align}
\end{small}
where the coefficients $a,b,c,d,e,f$ are found by solving the linear regression:
\begin{eqnarray}
\label{Eq:LSR}
\E{argmin}_{a,b,c,d,e,f}\frac{1}{T}\sum_{i=1}^T\|v_{i+1}^{out}-g(\theta_{i+1},vol_i)\|^2,
\end{eqnarray}
using a set of training data acquired from CFD simulation. One issue with this setting of coefficients is that $v^{out}$ can be negative, although this never happens in our experiments. In summary, our governing PDE $f$ is posed as:
\begin{eqnarray}
\label{Eq:Pred}
f(\E{p}_{i+1},\E{q}_i)=\TWOC{vol_{i+1}}{v_{i+1}^{out}}
=\TWOC{vol_i-A(\theta_{i+1},vol_i)v_{i+1}^{out}\Delta t}{g(\theta_{i+1},vol_i)}.
\end{eqnarray}
This simplified model works well if the source container $\E{S}$ is moving slowly and smoothly, which is consistent with the goal of the smoothness objective $C_{reg}$. See \prettyref{Sec:ER} for more discussion.

\subsection{LIQUID TRANSFER OBJECTIVE FUNCTION}
Given the governing PDE from \prettyref{Sec:GPDE}, we can now formulate our objective function $C_{obj}$. In principle, $C_{obj}$ requires the free-flying stage curve $\E{C}$ to pass through the center of the opening of the target container, denoted as $\E{O}_T$. We assume that the mean flow follows a quadratic curve, parametrized by:
\begin{equation}
\label{Eq:CurveC}
\E{C}(t)=\frac{g}{2}t^2+V_i^{out}t+E_i,
\end{equation}
and we penaltize the distance between $\E{C}(t)$ and $\E{O}_{\E{T}}$ when they are at the same altitude. Therefore, our objective function can be defined as:
\begin{eqnarray}
\label{Eq:CObj}
C_{obj}=\sum_iA(\theta_i,vol_i)\|\frac{g}{2}t^2+V_i^{out}t+E_i-\E{O}_{\E{T}}\|^2,
\end{eqnarray}
where $t$ is computed by solving $<g,\E{C}(t)-\E{O}_{\E{T}}>=0$, which is the time when the curve reaches the same altitude as $\E{O}_{\E{T}}$; $V_i^{out}$ maps $v_i^{out}$ in local frame of $\E{S}$ to the global frame and $E_i$ is the centroid of cross section area $A(\theta_i,vol_i)$. More details of their derivation are given in \prettyref{Appen:Geom}. Note that this term is weighted by the cross section area, so that it only takes effect when the outflow flux is nonzero. However, this objective alone is not sufficient because in our initial trajectory $A(\theta,vol)$ is zero in every timestep. As a result, we add a guiding term: $\|\theta_N-\theta_{final}\|^2$ to enforce a fixed final leaning angle. We use $\theta_{final}=90^\circ$ in all the experiments.
\section{\label{Sec:ER}EXPERIMENTS AND RESULTS}
In this section, we give details of our implementation and highlight the performance of our planner. We also evaluate the accuracy and efficacy of our simplified model.  

\begin{figure}[t]
\begin{center}
\includegraphics[trim=-100mm 0mm -100mm 0mm, clip, width=.99\linewidth]{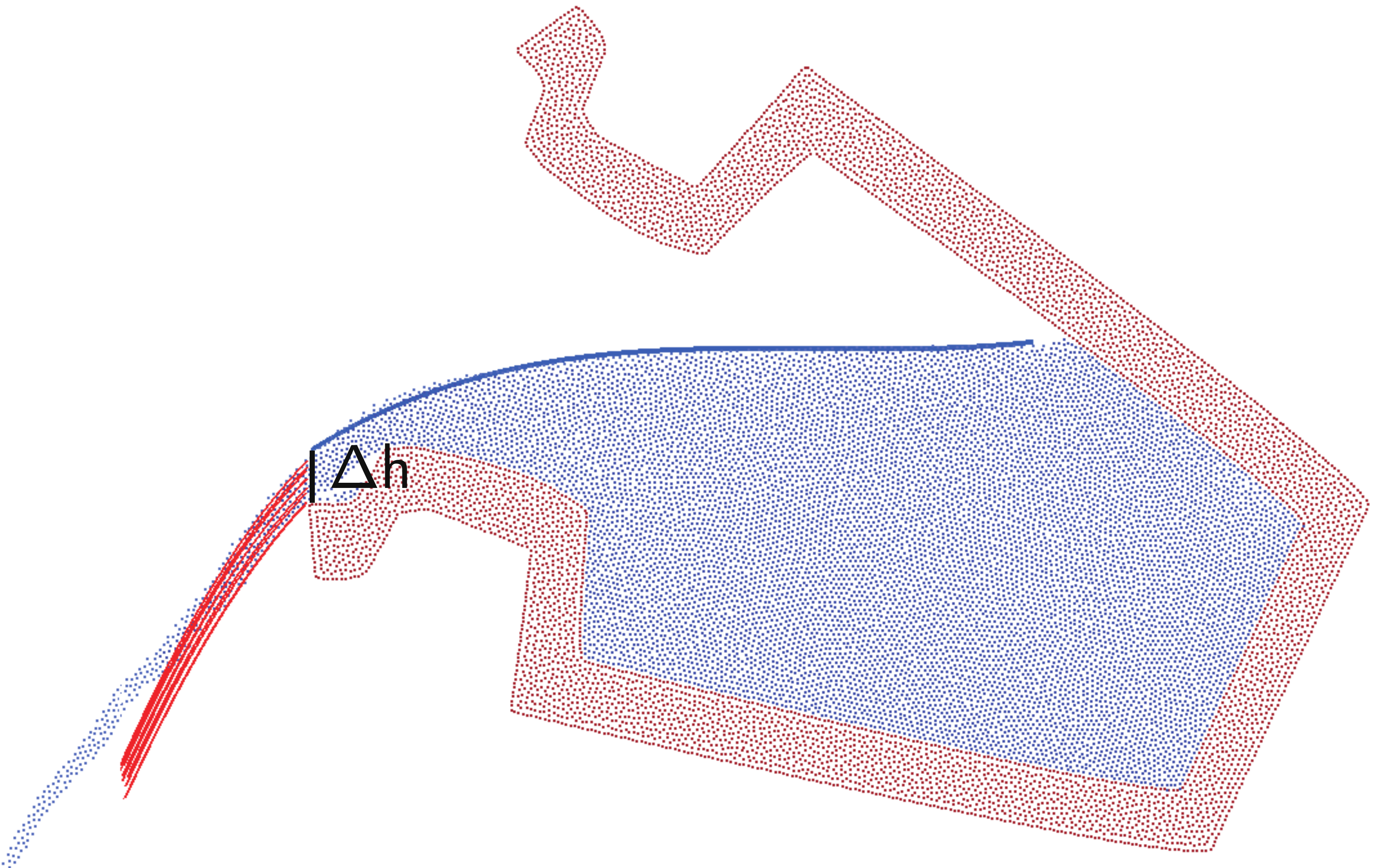}
\end{center}
\vspace{-10pt}
\caption{\label{Fig:Train} A snapshot of the cross section of fluid simulation, the extracted quadratic curves (red) for particles leaving $\E{S}$ at current timestep. Their tangents are averaged to compute $v^{out}$, and the cubic curve fitted to the free surface (blue). This is used to compute $\Delta h$ by evaluating the free surface position at boundary of $\E{S}$.}
\vspace{-10pt}
\end{figure}
In order to find the unknown coefficients involved in our simplified model, we need to solve the linear regression \prettyref{Eq:LSR} from a set of training data. The dataset is acquired from full-featured exact fluid simulations. Our simulator is implemented as a low order discretization of the Naiver-Stokes equation proposed in \cite{zhu2005animating,batty2007fast}. The fluid body in the simulator is represented by a set of particles as shown in \prettyref{Fig:FluidRep}. For each mesh of container $\E{S}$ and each fluid material setting, we need to solve for a separate set of coefficients $a,b,c,d,e,f$. In order to compute the training data, we ran 127 3D fluid simulations of pouring liquid out of a container by leaning it from $\theta=0^\circ$ to $\theta\in[90^\circ,150^\circ]$ using a random $\dot{\theta}$. This results in approximately 20000 training samples, each of which is a tuple $<v^{out},\theta,\Delta h>$. To extract these samples, for the set of particles in every timestep of the simulation, We extract their 2D axial symmetric cross-section. In this cross section, we compute $v^{out}$ by averaging the magnitude of velocity of all particles that leave the container $\E{S}$ at current timestep. These particles should be inside the bounding volume of $\E{S}$ in the previous timestep, but outside it in the current timestep. We also compute the characteristic height $\Delta h$, which is found by fitting a cubic curve to the free surface and evaluate it at the step boundary. A snapshot of the training data is illustrated in \prettyref{Fig:Train}.

\begin{figure}[t]
\begin{center}
\scalebox{1.0}
{
\includegraphics[trim=0mm 0mm -20mm 0mm, clip, width=.99\linewidth]{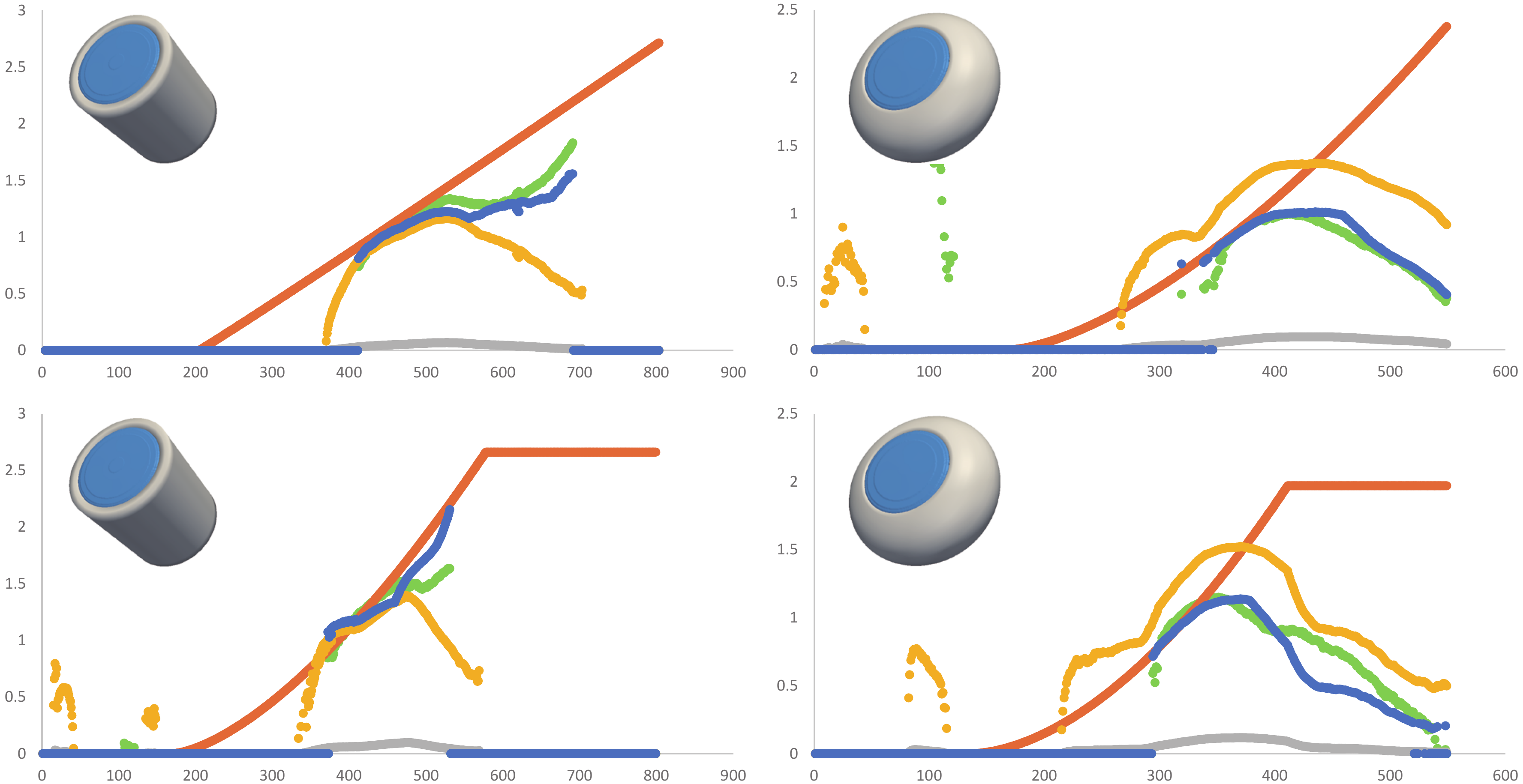}
\put (-245,125) {
\textcolor[rgb]{0.56,0.82,0.31}{\small$v^{out} (m/s)$}\;
\textcolor[rgb]{1,0.75,0}{\small$\sqrt{2gh} (m/s)$}\;
\textcolor[rgb]{0.93,0.49,0.19}{\small$\theta (rad)$}\;
\textcolor[rgb]{0.64,0.64,0.64}{\small$\Delta h (m)$}\;
\textcolor[rgb]{0.27,0.45,0.77}{\small$g(\theta,vol) (m/s)$}}
}
\put (-145,95) {(a)}
\put (-25,95) {(c)}
\put (-145,25) {(b)}
\put (-25,25) {(d)}
\end{center}
\vspace{-10pt}
\caption{\label{Fig:Error} For two simulated testing trajectories and two container shapes, we plot changes of the five variables $v^{out},\sqrt{2gh},\theta,\Delta h,g(\theta,vol)$ over time. Both testing and training trajectories are generated by simulating the liquid while turning the container according to a random, monotonically increasing $\theta$ curve shown in orange. The green curve is the groundtruth $v^{out}$, the outflowing velocity magnitude; the gray curve is the groundtruth $\Delta h$, and the yellow curve is the prediction of $v^{out}$ made by the Bernoulli equation $\sqrt{2g\Delta h}$; Finally, the blue curve is the predicted $v^{out}$ generated using our simplified dynamics model $g(\theta,vol)$. The error between $v^{out}$ and $g(\theta,vol)$ is small over the effective range where $v^{out}>0$.}
\end{figure}
\begin{figure}[t]
\begin{center}
\scalebox{1.0}
{
\includegraphics[trim=0mm 0mm -20mm 0mm, clip, width=.99\linewidth]{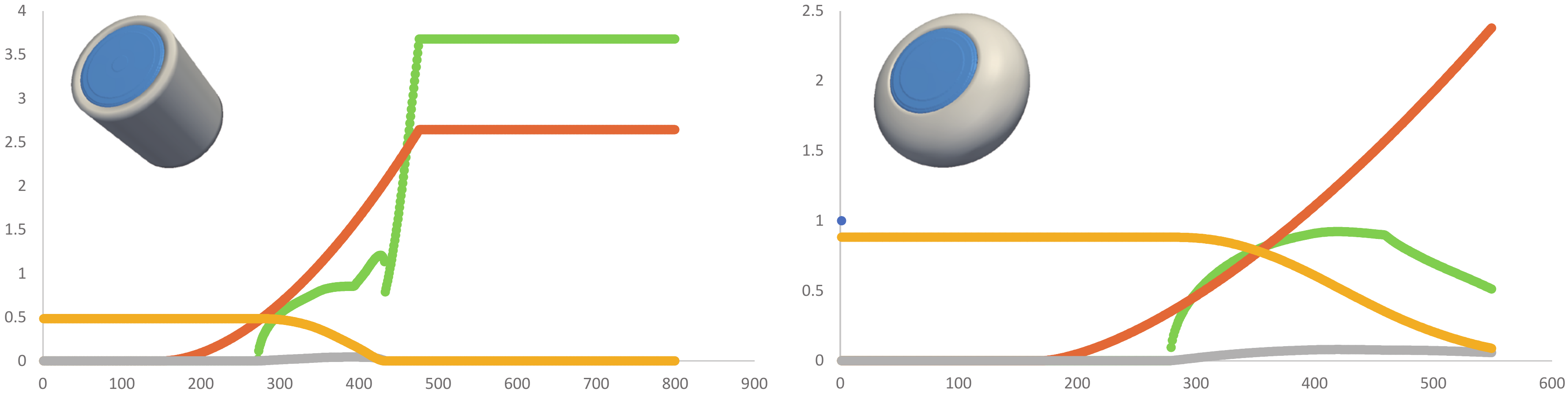}
\put (-245,65) {
\textcolor[rgb]{0.56,0.82,0.31}{\small$v^{out} (m/s)$}\;
\textcolor[rgb]{1,0.75,0}{\small$vol (m^3)$}\;
\textcolor[rgb]{0.93,0.49,0.19}{\small$\theta (rad)$}\;
\textcolor[rgb]{0.64,0.64,0.64}{\small$\Delta h (m)$}}
}
\end{center}
\vspace{-10pt}
\caption{\label{Fig:Synthesis} Predicted liquid trajectory $\E{P}$ computed using \prettyref{Eq:Pred}, for the two container shapes. Over time, the leaning angle $\theta$ increases monotonically (orange) while the total volume decreases accordingly (yellow). Note that the change of $v^{out}$ (green) closely resembles the groundtruth data (\prettyref{Fig:Error} green): for the cylindrical container, $v^{out}$ is always increasing while for the oval-shaped container, $v^{out}$ first increases and then decreases.}
\vspace{-10pt}
\end{figure}
After solving for the coefficients, we want to evaluate the accuracy of our simplified model. We plot the change of several related variables with respect to physical time in \prettyref{Fig:Error}. We note from the \prettyref{Fig:Error} (c,d) that, for the oval-shaped container, the Bernoulli approximation $\sqrt{2g\Delta h}$ can already achieve acceptable accuracy. However, for the cylindrical container, Bernoulli approximation leads to large error when $\theta\geq\pi/2$. In these cases, $\E{sin}(\E{max}(\theta-\pi/2))$ is a better approximation to $v^{out}$. By combining the six terms in \prettyref{Eq:Model}, we can achieve much better agreement with the groundtruth $v^{out}$. And by plugging the function $g$ into \prettyref{Eq:Pred}, we can even predicate the entire liquid trajectory, i.e. predicating $\E{P}$ given $\E{Q}$. Two of such predicted trajectories are illustrated in \prettyref{Fig:Synthesis}. Our model is flexible enough to account for different container shapes. Note that Bernoulli approximation can result in some false predictions when $\Delta h=0$ (and thus $A=0$), e.g. the yellow dots in the early timesteps of \prettyref{Fig:Error}, this will not cause any problems since our objective function in \prettyref{Eq:CObj} does not take into effect such degenerate cases.

\begin{table}
\begin{center}
\resizebox{\columnwidth}{!}{
\begin{tabular}{ccccc}
BENCHMARK & VISCOSITY & \shortstack{TIME\\PLANNING} & \shortstack{TIME\\VALIDATION} & QUALITY  \\
\hline  \\
\prettyref{Fig:Bench} (a) & $\mu=0.01$ & 5min & 2.1h & 95.7\%    \\
\prettyref{Fig:Bench} (a) & $\mu=0.001$ & 5min & 2.5h & 89.5\%   \\
\prettyref{Fig:Bench} (c) & $\mu=0.01$ & 7min & 1.9h & 93.2\%    \\
\prettyref{Fig:Bench} (c) & $\mu=0.001$ & 7min & 2.2h & 87.1\%
\end{tabular}
}
\end{center}
\caption{\label{Table:Profile} From left to right: benchmark environment, viscosity of fluid, time spent running \prettyref{Alg:pipeline}, time spent running forward CFD simulation for validation and quality of planned trajectory. The quality is measured by the fraction of particles that fall inside $\E{T}$.}
\vspace{-30pt}
\end{table}
To evaluate the motion planning pipeline, we plug the functions $f$ and $C_{obj}$ into \prettyref{Alg:pipeline}. This algorithm is implemented on ROS \cite{sucan2013moveit} platform. In order to implement the trust region SQP optimizer, we use the OOQP lib \cite{gertz2003object} to solve each QP subproblem. For collision constraints, although \cite{schulman2014motion} used $C_{collP}^1$, we use $C_{collP}^2$ in all our experiments because $C_{collP}^2$ introduces half the number of constraints in the QP problem. These constraints correspond to the box constraints that the OOQP algorithm can handle efficiently. Our simulated robot is a 9-DOF low cost ClamArm. All the experiments are setup on a single desktop computer with i7-4790 8-core CPU 3.6GHz and 8GB of memory. We evaluated the system on two liquid transfer tasks with different sets of obstacles in the environment. For each benchmark, we use two sets of fluid materials that differ only in their viscosity ($\mu=0.01,0.1$), so that two trained models are needed for each material and each source container $\E{S}$. We sample each trajectory with 100 nodes ($N=100$).

Several timesteps are illustrated in \prettyref{Fig:Bench}, to validate the quality of these trajectories. We replace the simplified dynamics model with a full-featured CFD simulator and our quality measure is the fraction of fluid particles that fall inside the target container $\E{T}$. These quality measures and time cost of planning are summarized in \prettyref{Table:Profile}. Each execution of \prettyref{Line:Inner} in \prettyref{Alg:pipeline} takes less than 5 seconds, while a full-featured CFD simulation in \cite{Zherong:2016:ICAPS} takes more than 2 hours. Thus we achieve at least two orders of magnitude speedup using our simplified dynamics model, as opposed to an exact fluid simulator and achieve comparable accuracy.

\section{CONCLUSION, LIMITATIONS, FUTURE WORK}
In this paper, we present an optimization based motion planning algorithm for liquid transfer. Our formulation uses an optimization-based planner and uses a simplified dynamics model to for the fluid constraint. The iterative procedure allows us to handle the non-smooth governing equation $f$, while the simplified dynamics model can be used to replace full-featured CFD simulation. This avoids introducing a large set of parameters to represent the fluid model, and thereby leading to a large speedup. We have evaluated the accuracy of our simplified model as well as the planning algorithm by comparing them with fluid simulation groundtruth data. The result shows that our model is able to predict the mainflow location when moving and leaning source container at mild speed.

The limitations of current work are due to various assumptions, such as the symmetry of the source container and the fixed position of target container. More importantly, our simplified parameter set and the dynamics model is restricted to the particular task of liquid transfer at mild speed. For example, when the end-effector is moving too fast, the predicated curve $\E{C}$ can be far from the groundtruth and the planned trajectory $\E{P}$ may not be physically correct.

In terms of future works, we would like to introduce additional constraints on the spacetime optimization formulation, such that our simplified model's prediction error is below some threshold. Another possible extension is to use more accurate, but still reduced parameter set and dynamics model, e.g. using convolutional neural network.
\section{\label{Appen:Geom} APPENDIX: THE CONTAINER GEOMETRY}
\begin{figure}[t]
\begin{center}
\includegraphics[trim=0mm 0mm 0mm 0mm, clip, width=.98\linewidth]{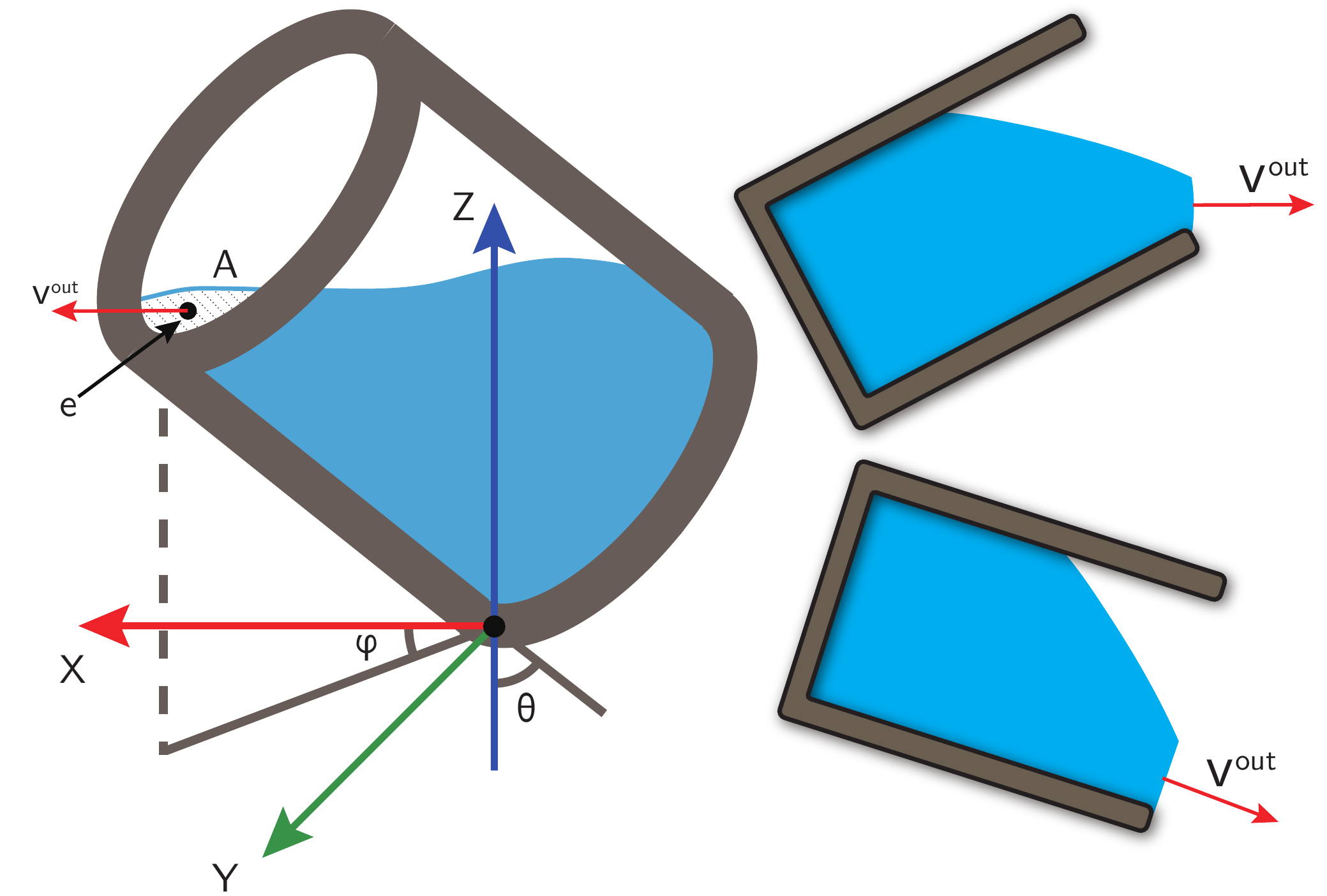}
\end{center}
\vspace{-10pt}
\caption{\label{Fig:Local} Left: An illustration of the orientation related parameter for source container $\E{S}$. The cross section area $A$ is the shaded region with centroid $e$. The outflowing velocity has magnitude $v^{out}$. Right: The outflow direction is horizontal when $\theta<90^\circ$. Otherwise, it is along the tangent direction.}
\vspace{-10pt}
\end{figure}

To characterize the quadratic curve $\E{C}$, we need the centroid $E_i$ of cross section area $A_i$. Also, we need the relationship between $V_i^{out}$ and $v_i^{out}$. Since we assume the source container $\E{S}$ to be axial symmetric, its orientation can be characterized by two parameters $\theta,\phi$ as illustrated in \prettyref{Fig:Local}. Given these parameters we have:
\begin{eqnarray*}
E_i&=&\MTTTHREE
{\E{cos}\phi}{-\E{sin}\phi}{0}
{\E{sin}\phi}{ \E{cos}\phi}{0}
{0}{0}{1}e_i    \\
V_i^{out}&=&\MTTTHREE
{\E{cos}\phi}{-\E{sin}\phi}{0}
{\E{sin}\phi}{ \E{cos}\phi}{0}
{0}{0}{1}
\THREEC{v_i^{out}\E{sin}[\E{max}(\theta,\frac{\pi}{2})]}{0}{v_i^{out}\E{cos}[\E{max}(\theta,\frac{\pi}{2})]},
\end{eqnarray*}
where $e_i$ is the centroid of $A_i$ in the local coordinates of $\E{S}$, which is again assumed to be a function $e(\theta,vol)$ precomputed in the same way as $A(\theta,vol)$. The $\E{max}$ operator in $V_i^{out}$ is critical to the accuracy of our dynamics model. When the leaning angle is less then $90^\circ$, we assume the outflow direction to be horizontal, see bottom \prettyref{Fig:Local}.
\bibliographystyle{IEEEtranS}
\bibliography{planning}
\end{document}